\documentclass[10pt,twocolumn,letterpaper]{article}

\usepackage[pagenumbers]{cvpr} 

\usepackage{graphicx}
\usepackage{amsmath}
\usepackage{amssymb}
\usepackage{booktabs}
\usepackage{pifont}  
\usepackage{colortbl} 
\usepackage{xcolor} 
\usepackage{algorithm}
\usepackage{algpseudocode}
\usepackage{makecell}
\usepackage{multirow}
\usepackage[pagebackref,breaklinks,colorlinks]{hyperref}

\usepackage[capitalize]{cleveref}
\crefname{section}{Sec.}{Secs.}
\Crefname{section}{Section}{Sections}
\Crefname{table}{Table}{Tables}
\crefname{table}{Tab.}{Tabs.}

\def\cvprPaperID{*****} 
\def\confName{CVPR}
\def\confYear{2025}

\begin{document}

\title{A Simple Detector with Frame Dynamics is a Strong Tracker}

\author{
Chenxu Peng$^{1}$,
Chenxu Wang$^{1}$,
Minrui Zou$^{1}$,
Danyang Li$^{1}$,
Zhengpeng Yang$^{1}$,\\
Yimian Dai$^{1,2}$,
Ming-Ming Cheng$^{1,2}$,
Xiang Li$^{1,2}$\thanks{Corresponding author: Xiang Li.}\\
$^1$ VCIP, CS, Nankai University
$^2$ NKIARI, Shenzhen Futian\\
{\tt\small pcx0521@163.com},
{\tt\small facias914@gmail.com},
{\tt\small traveler\_wood@163.com},\\
{\tt\small \{2120240690, 2211819\}@mail.nankai.edu.cn},
{\tt\small yimian.dai@gmail.com},\\
{\tt\small \{cmm, xiang.li.implus\}@nankai.edu.cn}
}

\maketitle
\begin{abstract}
   Infrared object tracking plays a crucial role in Anti-Unmanned Aerial Vehicle (Anti-UAV) applications. Existing trackers often depend on cropped template regions and have limited motion modeling capabilities, which pose challenges when dealing with tiny targets. To address this, we propose a simple yet effective infrared tiny-object tracker that enhances tracking performance by integrating global detection and motion-aware learning with temporal priors. Our method is based on object detection and achieves significant improvements through two key innovations. First, we introduce frame dynamics, leveraging frame difference and optical flow to encode both prior target features and motion characteristics at the input level, enabling the model to better distinguish the target from background clutter. Second, we propose a trajectory constraint filtering strategy in the post-processing stage, utilizing spatio-temporal priors to suppress false positives and enhance tracking robustness.  Extensive experiments show that our method consistently outperforms existing approaches across multiple metrics in challenging infrared UAV tracking scenarios. Notably, we achieve state-of-the-art performance in the 4th Anti-UAV Challenge, securing 1st place in Track 1 and 2nd place in Track 2. The code are
publicly available at: \url{https://github.com/facias914/A-Simple-Detector-is-a-Strong-Tracker}.
\end{abstract}

\section{Introduction}
\label{Introduction}

\begin{figure}[t]
    \centering
    \includegraphics[width=1.0\linewidth]{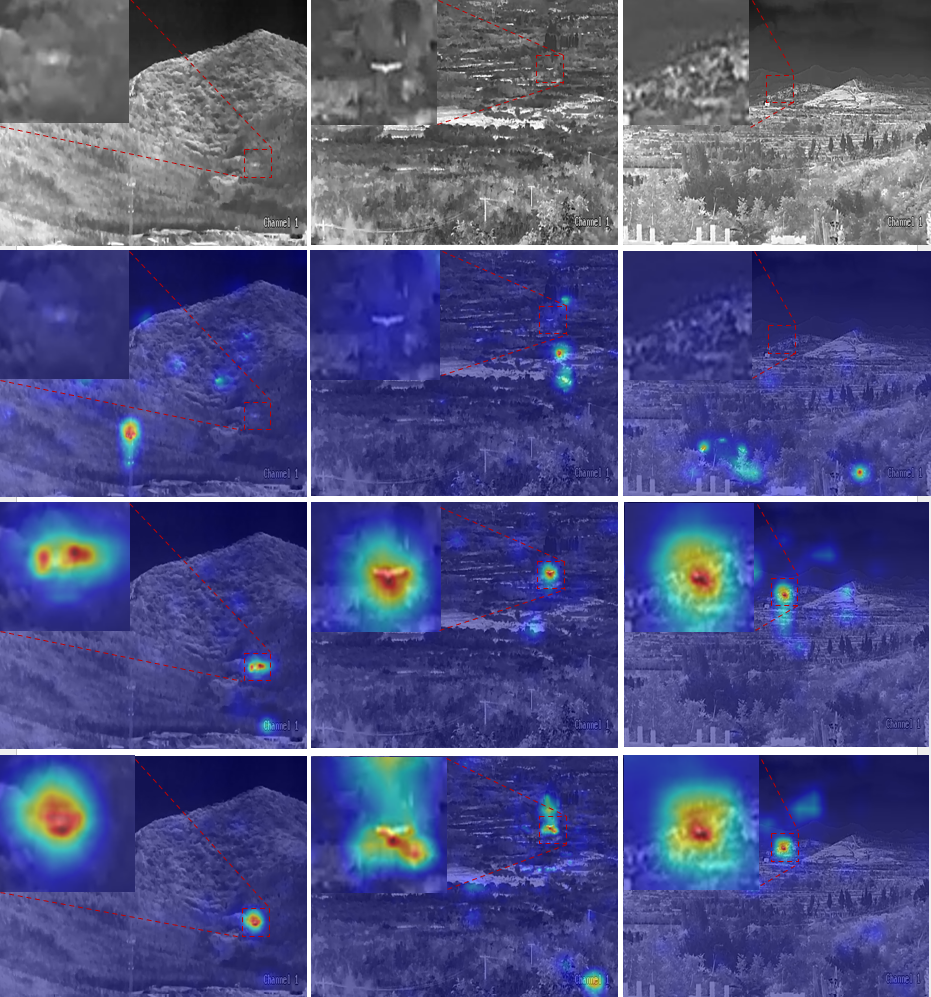} 
    \caption{Visualization of heatmaps from different detection models. The first row shows the original images, while the second to fourth rows correspond to the detection results of different types of model inputs: the original image, the original image concatenated with the frame difference map, and the original image concatenated with the optical flow map. The heatmaps are generated using the HiResCAM~\cite{draelos2020use} method.}
    \label{fig:heatmap}
\end{figure}

Object tracking~\cite{he2018twofold} is an extension of object detection~\cite{girshick2015fast, ren2015faster, lin2017focal}. It enables the continuous localization of specific targets in video sequences by leveraging template frames and temporal cues. Unlike traditional object detection, object tracking does not perform independent detection in each frame. This technique is essential in Anti-UAV applications~\cite{jiang2021anti, huang2023anti}. In this work, we focus on single-UAV tracking in infrared scenarios and propose a simple yet efficient infrared tiny-object tracker.

\begin{figure*}[htbp]
    \centering
    \includegraphics[width=\textwidth]{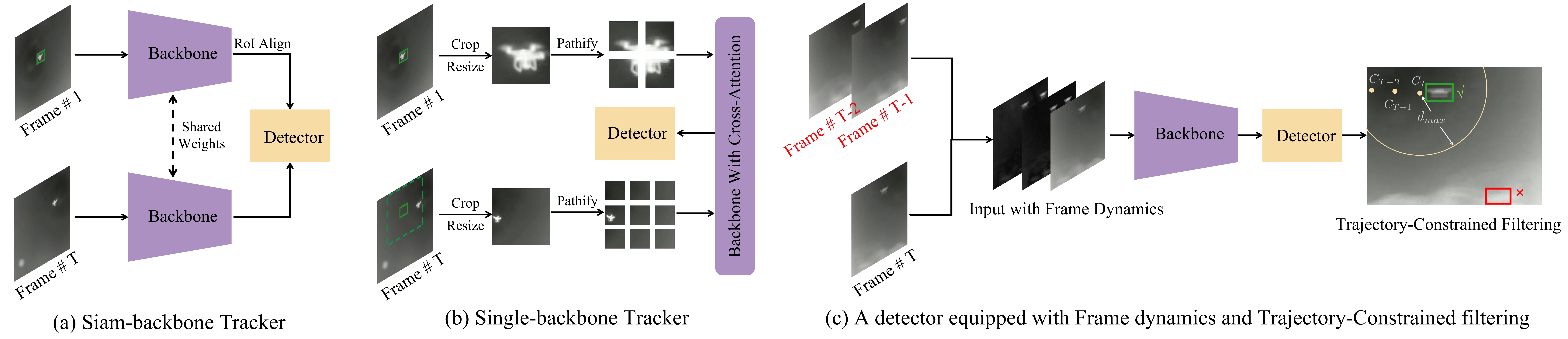} 
    \caption{Comparison between the conventional object trackers and the proposed object tracker. Figure (c) only utilizes the frame difference method as an example, while the optical flow method follows the same approach.}
    \label{fig:pipeline}
\end{figure*}

Object trackers are primarily built upon object detectors. Following this research direction, advanced trackers have evolved from two-stream architectures (Fig.~\ref{fig:pipeline}a ~\cite{he2018twofold, voigtlaender2020siam, huang2020globaltrack} to unified single-backbone networks (Fig.~\ref{fig:pipeline}b) \cite{ye2022joint, cui2022mixformer, lin2022swintrack, fan2024querytrack, yan2021learning}, driven by cross-attention mechanisms\cite{vaswani2017attention}. This transition has significantly enhanced the efficiency of feature extraction and utilization from template frames. However, the conventional approach of cropping and resizing the target region from the template frame is inadequate for scenarios involving tiny targets or cases where the target’s initial position is unknown (e.g., Track 2).

Furthermore, existing methods~\cite{ye2022joint,lin2022swintrack, fan2024querytrack} leverage temporal information during the data preprocessing stage (Fig.~\ref{fig:pipeline} (b)) by cropping the current frame based on the target's position and size in either the template or the previous frame. However, this strategy introduces several inherent limitations. First, the lack of effective background suppression leads the model to focus on the target itself, failing to adequately distinguish it from surrounding non-target regions. Second, this approach does not facilitate the learning of the target’s dynamic motion patterns, thereby limiting the model’s ability to capture temporal dependencies effectively. Third, false positive detections from the previous frame can propagate through subsequent frames during inference, leading to a cascading effect of accumulated misdetections, which severely impacts tracking robustness and accuracy.

Based on the above analysis, we believe that a robust infrared target tracker should satisfy the following conditions: 1) strong global detection capability, which can be achieved through object detecter; 2) the ability to learn not only the target's features from previous frames but also its motion characteristics; 3) effective utilization of temporal priors at appropriate times. Therefore, we first experimentally select several object detectors that perform excellently in infrared small target scenarios. Building on this foundation, we design the frame dynamics data using frame difference~\cite{wren1997pfinder} and optical flow methods~\cite{horn1981determining}, allowing the model to flexibly learn both the prior target features and the motion characteristics of previous frames at the input level. With the assistance of Frame Dynamics, the model can effectively focus on the specified target's features even in complex scenarios, as shown in Fig~\ref{fig:heatmap}. Furthermore, a trajectory constraint filtering strategy is introduced in the post-processing stage during inference, leveraging spatio-temporal priors to robustly filter false positives. By integrating these strategies, we successfully transform a simple object detector into a powerful object tracker.

\section{Related Work}

\subsection{Object Detection}

Infrared UAV detection relies on infrared radiation to identify targets but faces challenges such as low signal-to-noise ratio, poor contrast, and complex noisy backgrounds~\cite{he2023motion, xiao2024background}. In recent years, deep learning has excelled in this field by leveraging large-scale, diverse datasets~\cite{dai2023one, dai2024pick, yuan2024sctransnet, li2024lsknet}. Cascade R-CNN~\cite{cai2018cascade} employed a multi-stage refinement strategy with increasing IoU thresholds, enhancing localization accuracy and reducing false positives. DINO~\cite{zhang2022dino} utilized a Transformer-based architecture with contrastive denoising training and dynamic query selection, improving small object feature extraction in cluttered scenes. RepPoints~\cite{yang2019reppoints} modeled objects as learnable keypoints, enabling finer spatial representation and precise localization of irregularly shaped targets. PAA~\cite{kim2020probabilistic} introduced probabilistic anchor assignment, dynamically selecting high-quality anchors to improve detection robustness, particularly for densely packed small objects. These methods effectively addressed key challenges in small object detection, significantly enhancing detection performance in complex infrared imaging scenarios.

\subsection{Object Tracking}
Object tracking is a continuation of detection. Early works like TransT~\cite{chen2021transformer} and Stark~\cite{yan2021learning} enhanced feature extraction and fusion, while one-stream frameworks (e.g., MixFormer~\cite{cui2022mixformer}, SimTrack~\cite{chen2022backbone}, OSTrack~\cite{ye2022joint}) enabled joint template-search interaction. Recent studies (e.g., GRM~\cite{gao2023generalized}, DropTrack~\cite{wu2023dropmae}, SeqTrack~\cite{chen2023seqtrack}]) further improved relational modeling with adaptive token partitioning and autoregressive decoders. To improve small target tracking performance, SiamDT~\cite{huang2023anti} employed a dual-semantic feature mechanism, integrating Dual-Semantic RPN Proposals, Versatile R-CNN, and background distractor suppression to enhance tracking. In this work, we propose a simple yet effective infrared small object tracker that enhances tracking performance by integrating global detection, motion-aware learning, and temporal prior utilization.


\section{Evolving a Detector into a Tracker}
\label{Method}

In this section, we focus on transforming a common detector into a robust tracker for UAV tracking. Our approach is guided by two key distinctions (Fig.~\ref{fig:pipeline}c): 1) the fundamental differences between detection and tracking tasks, and 2) the unique challenges of infrared tiny object tracking compared to conventional scene tracking.

\subsection{Selecting a base detector}

Our starting point is selecting a detector as the base model.
The primary criterion for this selection is its strong performance in infrared tiny object detection scenarios.
Through experiments on different types of detectors, we identified five promising candidates: YOLOv11-customized (with $p2$ detector)~\cite{khanam2024yolov11}, Cascade R-CNN~\cite{cai2018cascade}, DINO~\cite{zhang2022dino}, RepPoints~\cite{yang2019reppoints}, and PAA~\cite{kim2020probabilistic}. Each model has been modified compared to its original version, and the specific adjustments will be detailed in the next section.

\subsection{Empowering detector with frame dynamics}

\begin{figure}[t]
    \centering
    \includegraphics[width=1.0\linewidth]{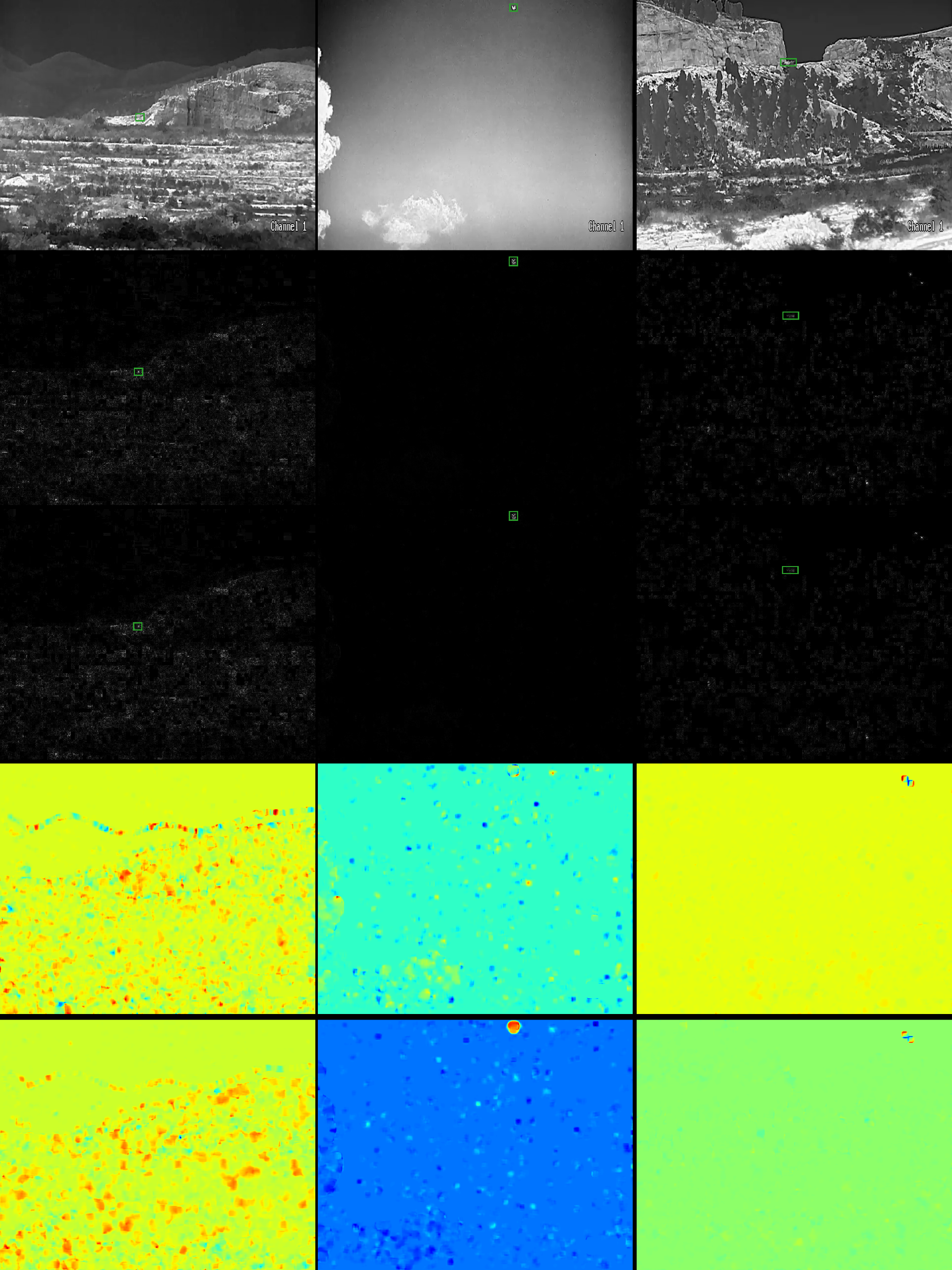} 
    \caption{Visualization of frame difference maps and optical flow maps. The first row shows the original images. The second and third rows display the frame difference maps between the previous frame and the current frame, as well as between the frame before last and the current frame, respectively. The fourth and fifth rows present the optical flow maps in the horizontal and vertical directions between the previous frame and the current frame. Each column represents a different scene. Note that the optical flow maps are actually grayscale images, but colors have been added here for better visualization.}
    \label{fig:df_of_vis_resized}
\end{figure}

The fundamental distinction between object tracking and object detection lies in object tracking’s ability to exploit appearance consistency and motion predictability beyond individual frames, which are
1) the appearance and feature representations of a target exhibit strong consistency across adjacent frames or throughout the entire sequence;
2) the target’s movement adheres to physical constraints, allowing its velocity, acceleration, and inertial properties to be inferred from previous frames, enabling precise position prediction.
Inspired by these, we propose a simple yet effective method for infrared small target tracking, where the lack of rich texture and color information in single-channel infrared images makes it challenging to distinguish targets based on individual frames alone. To fully leverage the appearance consistency and motion predictability priors, frame dynamics information is concatenated with the current frame as input to the detection network. Specifically, two methods are introduced for constructing frame dynamics, as shown in Fig.~\ref{fig:df_of_vis_resized}: the frame difference method and the optical flow method. Given the current frame $x_t$, the frame difference method constructs the input as:
\begin{equation}
x_{fd} = cat(x_t, x_t - x_{t-1}, x_t - x_{t-2}).
\end{equation}
We select the two preceding frames of the current frame to construct frame difference features, enabling the input data to capture short-term motion and feature information of the target.
As for the optical flow method, we employ the Farneback~\cite{vihlman2020optical} optical flow method to construct the optical flow between the current frame and the previous frame, as it enables the estimation of a global motion field, helping the model learn distinctive motion features relative to the background.
The input is constructed as:
\begin{equation}
x_{of} = cat(x_t, f(x_t, x_{t-1})_v, f(x_t, x_{t-1})_u),
\end{equation}
where $u$ and $v$ are the horizontal and vertical components respectively, and $f$ represents the optical flow function.
This formulation injects temporal information as a third dimension into the originally two-dimensional infrared images, allowing the model to capture both the target’s historical appearance features and motion dynamics, effectively transforming a detector into a strong tracker.

\subsection{Refining tracking with TC-Filtering}

Another important prior in video-based object tracking is that the target’s position in the current frame is constrained within a localized region around its absolute position in the previous frame, following the natural continuity of motion.
Therefore, we design a Trajectory-Constrained Filtering (TC-Filtering) strategy that estimates the target's motion trend based on the bounding boxes from the previous two frames and eliminates candidate boxes that deviate from the expected motion pattern.
Concretely, given the center points of the target bounding boxes in the previous two frames, $C_{t-2}=(x_{t-2}, y_{t-2})$ and $C_{t-1}=(x_{t-1}, y_{t-1})$, the instantaneous velocity of the target can be computed as:
\begin{equation}
\mathbf{V}_{t-1} = \begin{bmatrix} V_x \\ V_y \end{bmatrix} = \frac{1}{\Delta t} \begin{bmatrix} x_{t-1} - x_{t-2} \\ y_{t-1} - y_{t-2} \end{bmatrix},
\end{equation}
where $\Delta t$ is the time interval between two consecutive frames. 
Therefore, the center position of the target in the current frame can be approximately calculated as:
\begin{equation}
C_t \approx C_{t-1} + \mathbf{V}_{t-1} \cdot \Delta t 
= \begin{bmatrix} 2x_{t-1} - x_{t-2} \\ 2y_{t-1} - y_{t-2} \end{bmatrix}.
\end{equation}
Then a motion constraint window is defined with $C_t$ as the center and $d_{max}$ as the radius.
The center of the candidate box detected in the current frame must fall within the motion constraint window, otherwise it will be discarded.
\begin{equation}
\label{eq5}
S_t^{(i)} = 
\begin{cases} 
1 & \text{if } \left\| C_t^{(i)} - C_t \right\|_2 \leq d_{\text{max}} \\
0 & \text{otherwise}
\end{cases}.
\end{equation}
TC-Filtering is incorporated into the post-processing stage of the object detection inference pipeline, effectively leveraging the spatiotemporal continuity prior to constrain the target within a defined region, thereby substantially enhancing detection precision.

\section{Experiments}
\label{sec:formatting}

\subsection{Implementation Details}

{\bf Dataset Details.} The 4th Anti-UAV dataset\footnote{https://anti-uav.github.io/} is meticulously designed to facilitate the discovery, detection, recognition, and tracking of UAV targets in challenging environments. Additionally, it provides target tracking state estimation in thermal infrared videos, enhancing its utility for advanced UAV perception tasks. This dataset serves as the foundation for the Anti-UAV Challenge, a competition dedicated to advancing research and development in UAV detection and tracking algorithms under complex and dynamic conditions. It comprises three subsets: the training set, the test set for Track 1, and the test set for Track 2. The training set includes 223 thermal infrared video sequences, each enriched with detailed annotations such as target presence, target location, and various environmental labels. Each video frame features a resolution of 640 × 512 or 512 × 512, with individual videos containing up to 1,500 frames.

{\bf Evaluation Metrics.} For detection model evaluation, we use Average Precision (AP) as the core metric. However, AP alone cannot fully capture tracking stability in long sequences. Thus, we incorporate Average Overlap Accuracy (AOA)~\cite{zhao20233rd}, which measures the average IoU across all frames, to assess tracking consistency.
While AP evaluates detection accuracy, AOA reflects temporal stability, making them complementary. To enhance rigor, the 3rd Anti-UAV Challenge introduced a penalty term in AOA to penalize missed tracks when the target is present, ensuring a more realistic robustness assessment. More details can be found on the 3rd Anti-UAV Challenge page.

{\bf Dataset Processing and Split.} After reviewing the raw data, we found that approximately 5\% of the annotations contained errors. The primary issues included bounding boxes that did not align with the actual targets and instances where the bounding boxes were incorrectly assigned to the entire image. We have uploaded the download link for the corrected training dataset to our GitHub. Training with the revised dataset yields an improvement of 1.5 AP. 

\begin{algorithm}[t]
    \caption{\textbf{Four-Fold Cross-Validation with Similarity Constraints}}
    \label{tab:algorithm_cv}
    \begin{algorithmic}[1]
        \State \textbf{Input:} Image set $I$, Similarity dictionary $S$, Threshold $T$
        
        \State \textbf{Step 1: Construct Similarity Graph} $G$
        \For{each image $I_i \in I$}
            \For{each $I_j$ with $S(I_i, I_j) > T$}
                \State Compute Euclidean distance:
                \[
                d(I_i, I_j) = \sqrt{(x_i - x_j)^2 + (y_i - y_j)^2}
                \]
                \If{$d(I_i, I_j) < d_v$}
                    \State Connect $I_i \leftrightarrow I_j$ in $G$
                \EndIf
            \EndFor
        \EndFor
        
        \State \textbf{Step 2: Identify Image Groups via DFS}
        
        \State \textbf{Step 3: Sort Groups by Number of Images}
        \[
        G' = \{ G_i \mid G_i \in G, |G_1| \geq |G_2| \geq \dots \geq |G_n| \}
        \]

        \State \textbf{Step 4: Assign Groups to 4 Folds (Greedy Strategy)}
        \[
        \min \max \sum_{i=1}^{4} |F_i|
        \]

        \State \textbf{Step 5: Compute Validation Set Imbalance}
        \[
        \Delta_{val} = \max |V_i| - \min |V_i|
        \]
        
        \State \textbf{Output:} Training and validation splits
    \end{algorithmic}
\end{algorithm}

\begin{table}[t]
    \centering
    \caption{Main results of the 4th Anti-UAV workshop track-1 and track-2.}
    \label{tab:main_results}
    \definecolor{Gray}{gray}{0.9} 
    \resizebox{0.39\textwidth}{!}{  
    \begin{tabular}{ccc}
        \toprule
        \multicolumn{3}{c}{\textbf{Track-1}} \\
        \midrule
        Rank & Team Name & Tracking Performance (AOA) \\
        \midrule
        \cellcolor{Gray}1 & \cellcolor{Gray}Ours & \cellcolor{Gray}\textbf{73.23} \\
        2 & Anonymous & 73.08 \\
        3 & Anonymous & 71.45 \\
        \midrule
        \multicolumn{3}{c}{\textbf{Track-2}} \\
        \midrule
        Rank & Team Name & Tracking Performance (AOA) \\
        \midrule
        1 & Anonymous & \textbf{66.76} \\
        \cellcolor{Gray}2 & \cellcolor{Gray}Ours & \cellcolor{Gray}57.12 \\
        3 & Anonymous & 55.30 \\
        \bottomrule
    \end{tabular}
    }
\end{table}

Constructing a robust offline validation set is crucial to ensuring that evaluation metrics align with those of the online test set, thereby improving model generalization. To minimize data leakage, we design a structured validation set construction process, as outlined in Algorithm~\ref{tab:algorithm_cv}. First, we uniformly sample four frames from each video sequence and concatenate them into a single composite image to preserve spatial and temporal information. Next, we perform feature matching across composite images from different sequences to detect potential data overlap. Specifically, we employ the $SuperPoint + LightGlue$~\cite{detone2018superpoint, lindenberger2023lightglue} matching method, setting a threshold of 100 keypoint matches. If two composite images exceed this threshold, they are deemed highly similar and assigned to the same dataset (either training or validation) to prevent leakage. To further ensure data integrity, we apply a depth first search (DFS) algorithm to construct a four-fold cross-validation set, eliminating potential leakage between different folds. This structured approach enhances the reliability of the validation set, ensuring it provides an accurate assessment of model performance in real-world testing scenarios.

{\bf Training Details.} To reduce data redundancy, we applied a one-in-five sampling strategy for images in each video sequence within the training set. The validation set remained unchanged. Experimental results indicated that sampling, combined with more training epochs, yielded better performance. In contrast, training without sampling but with fewer epochs resulted in lower accuracy. A detailed experimental analysis is provided in the next section.

For YOLO model parameter settings, we designed and trained three models, each corresponding to a distinct input format: raw images, concatenated raw images with frame difference maps, and concatenated raw images with optical flow maps. To optimize performance, we developed customized data preprocessing and training strategies for each input type. During data preprocessing, input images were standardized by resizing the longest side to 640 pixels, with random scaling applied in the range of 0.7 to 1.3 for data augmentation. Images were normalized to zero mean and unit variance. Additionally, Mixup with a probability of 0.1 and random erasing with a probability of 0.4 were incorporated to enhance generalization. For the concatenated “Raw Image + Frame Difference” and “Raw Image + Optical Flow” inputs, HSV augmentation was disabled to preserve the integrity of temporal information. In terms of model architecture, YOLOv11x was selected as the backbone. The decoder was modified by integrating an Asymptotic Feature Pyramid Network (AFPN)\cite{yang2023afpn} to enhance multiscale feature fusion. A $p2$ detection head was introduced to improve small-object detection capability, while the C3K2\cite{alif2024yolov11} module was incorporated to optimize feature representation. For training, the SGD optimizer was used with a batch size of 32 over 300 epochs. To improve model stability and optimize final performance, Mosaic augmentation was disabled during the last five epochs. All other hyperparameters followed the default YOLOv11x settings.

\begin{table}[t]
    \centering
    \caption{Detection performance comparison (AP@50). The models highlighted in the gray area are the ones adopted in the final scheme.}
    \label{tab:map50_comparison}
    \definecolor{Gray}{gray}{0.9} 
    \resizebox{0.48\textwidth}{!}{ 
    \begin{tabular}{c c c}
        \toprule
        \textbf{Framework} & \textbf{Model} & \textbf{AP@50} \\
        \midrule
        \multirow{14}{*}{MMDetection} 
        & DETR~\cite{zhu2020deformable} & 0.1 \\
        & RetinaNet~\cite{lin2017focal} & 22.0 \\
        & Faster R-CNN~\cite{girshick2015fast} & 26.2 \\
        & Libra R-CNN~\cite{pang2019libra} & 28.4 \\
        & ConvNeXt + Faster R-CNN~\cite{woo2023convnext} & 28.8 \\
        & Dynamic R-CNN~\cite{zhang2020dynamic} & 31.3 \\
        & FCOS~\cite{tian2019fcos} & 37.0 \\
        & PVT~\cite{wang2022pvt} & 37.6 \\
        & ATSS~\cite{zhang2020bridging} & 38.3 \\
        & VFNet~\cite{zhang2021varifocalnet} & 39.1 \\
        & CenterNet~\cite{duan2019centernet} & 42.4 \\
        & TOOD~\cite{feng2021tood} & 43.2 \\
        & GFL~\cite{li2020generalized} & 44.3 \\
        & AutoAssign~\cite{zhu2020autoassign} & 45.6 \\
        \rowcolor{Gray} & Cascade R-CNN~\cite{cai2018cascade} & \textbf{52.4} \\
        \rowcolor{Gray} & PAA~\cite{kim2020probabilistic} & \textbf{52.5} \\
        \rowcolor{Gray} & RepPoints~\cite{yang2019reppoints} & \textbf{52.9} \\
        \rowcolor{Gray} & DINO~\cite{zhang2022dino} & \textbf{54.9} \\
        \midrule
        \multirow{7}{*}{YOLOv11} 
        & YOLO11n-slimneck~\cite{li2022slim} & 48.1 \\
        & YOLO11n-SPDConv~\cite{spd-conv2022} & 49.1 \\
        & YOLO11n~\cite{khanam2024yolov11} & 50.5 \\
        & YOLO11n-ASF-P2345~\cite{kang2024asf} & 51.9 \\
        & YOLO11n-ReCalibrationFPN-P2345~\cite{yang2025mhaf} & 52.5 \\
        & YOLO11n-SwinTiny-AFPN-P234~\cite{liu2021swin} & 54.0 \\
        \rowcolor{Gray} & YOLO11n-AFPN-P2345-C3K2h~\cite{liu2018path} & \textbf{55.0} \\
        \bottomrule
    \end{tabular}
    }
\end{table}

\begin{table*}[t]
    \centering
    \caption{Detection and tracking performance metrics of five models on the training set of the fourth competition. "Raw" refers to the original image, "Raw+FD" represents the concatenation of the original image and the frame differences of the previous two frames, while "Raw+OF" denotes the concatenation of the original image with the horizontal and vertical optical flow of the previous frame.}
    \label{tab:method_comparison}

    \definecolor{Gray}{gray}{0.9}

    \resizebox{\textwidth}{!}{
    \begin{tabular}{c|c|c|c|cccc|cccc}
        \toprule
        \multirow{3}{*}{\textbf{Methods}} & 
        \multirow{3}{*}{\textbf{Backbone}} & 
        \multirow{3}{*}{\textbf{Decoder}} & 
        \multirow{3}{*}{\textbf{Image Size}} &
        \multicolumn{4}{c|}{\textbf{Fold1}} &
        \multicolumn{4}{c}{\textbf{Fold2}} \\
        \cmidrule(lr){5-8} \cmidrule(lr){9-12}
        & & & &
        \makecell{\textbf{Raw}} & 
        \makecell{\textbf{Raw} \\ \textbf{+FD}} & 
        \makecell{\textbf{Raw} \\ \textbf{+OF}} & 
        \makecell{\textbf{Mean}} &
        \makecell{\textbf{Raw}} & 
        \makecell{\textbf{Raw} \\ \textbf{+FD}} & 
        \makecell{\textbf{Raw} \\ \textbf{+OF}} & 
        \makecell{\textbf{Mean}} \\
        \midrule
        \multicolumn{12}{c}{\textbf{Detection Performance (AP@50)}} \\
        \midrule
        Cascade R-CNN       & ResNet101 & PAFPN & 1280×1024 & 56.2 & 63.3 & 60.5 & 60.0 & 51.9 & 56.0 & 53.3 & 53.7 \\
        DINO                & ResNet101 & PAFPN & 1280×1024 & 57.6 & 65.9 & 62.7 & 62.1 & 51.6 & 57.7 & 55.6 & 55.0 \\
        RepPoints           & ResNet101 & PAFPN & 1280×1024 & 59.6 & 67.3 & 63.0 & 63.3 & 54.2 & 60.5 & 54.9 & 56.5 \\
        PAA                 & ResNet101 & PAFPN & 1280×1024 & 59.6 & \textbf{68.9} & \textbf{66.9} & \textbf{65.1} & 54.9 & 61.2 & 57.8 & 58.0 \\
        YOLOv11x-customized & YOLOv11x & AFPN & 640×512 & \textbf{60.5} & 63.8 & 63.1 & 62.5 & \textbf{55.7} & \textbf{64.0} & \textbf{61.2} & \textbf{60.3} \\
        \midrule
        \multicolumn{12}{c}{\textbf{Tracking Performance (AOA)}} \\
        \midrule
        Tracking WBF & --- & --- & --- 
            & \multicolumn{4}{c|}{46.3} 
            & \multicolumn{4}{c}{40.5} \\
        \cellcolor{Gray}Tracking WBF + TC-filtering & \cellcolor{Gray}--- & \cellcolor{Gray}--- & \cellcolor{Gray}--- 
            & \multicolumn{4}{c|}{\cellcolor{Gray}\textbf{48.1}} 
            & \multicolumn{4}{c}{\cellcolor{Gray}\textbf{42.8}} \\
        \bottomrule
    \end{tabular}
    }

    \vspace{-0.5em}

    \resizebox{\textwidth}{!}{
    \begin{tabular}{c|c|c|c|cccc|cccc}
        \toprule
        \multirow{3}{*}{\textbf{Methods}} & 
        \multirow{3}{*}{\textbf{Backbone}} & 
        \multirow{3}{*}{\textbf{Decoder}} & 
        \multirow{3}{*}{\textbf{Image Size}} &
        \multicolumn{4}{c|}{\textbf{Fold3}} &
        \multicolumn{4}{c}{\textbf{Fold4}} \\
        \cmidrule(lr){5-8} \cmidrule(lr){9-12}
        & & & &
        \makecell{\textbf{Raw}} & 
        \makecell{\textbf{Raw} \\ \textbf{+FD}} & 
        \makecell{\textbf{Raw} \\ \textbf{+OF}} & 
        \makecell{\textbf{Mean}} &
        \makecell{\textbf{Raw}} & 
        \makecell{\textbf{Raw} \\ \textbf{+FD}} & 
        \makecell{\textbf{Raw} \\ \textbf{+OF}} & 
        \makecell{\textbf{Mean}} \\
        \midrule
        \multicolumn{12}{c}{\textbf{Detection Performance (AP@50)}} \\
        \midrule
        Cascade R-CNN       & ResNet101 & PAFPN & 1280×1024 & 80.2 & 83.2 & 80.9 & 81.4 & 60.7 & 67.8 & 62.9 & 63.8 \\
        DINO                & ResNet101 & PAFPN & 1280×1024 & 75.6 & 81.2 & 78.3 & 78.4 & 62.9 & 68.4 & 65.8 & 65.7 \\
        RepPoints           & ResNet101 & PAFPN & 1280×1024 & 80.1 & 81.8 & 80.1 & 80.7 & 64.1 & 68.7 & 66.0 & 66.3 \\
        PAA                 & ResNet101 & PAFPN & 1280×1024 & 81.5 & 81.7 & 81.3 & 81.5 & 62.2 & 69.0 & 66.7 & 66.0 \\
        YOLOv11x-customized & YOLOv11x & AFPN & 640×512 & \textbf{82.5} & \textbf{84.6} & \textbf{83.1} & \textbf{83.4} & \textbf{67.0} & \textbf{72.7} & \textbf{70.1} & \textbf{70.0} \\
        \midrule
        \multicolumn{12}{c}{\textbf{Tracking Performance (AOA)}} \\
        \midrule
        Tracking WBF & --- & --- & --- 
            & \multicolumn{4}{c|}{77.1} 
            & \multicolumn{4}{c}{49.6} \\
        \cellcolor{Gray}Tracking WBF + TC-filtering & \cellcolor{Gray}--- & \cellcolor{Gray}--- & \cellcolor{Gray}--- 
            & \multicolumn{4}{c|}{\cellcolor{Gray}\textbf{80.8}} 
            & \multicolumn{4}{c}{\cellcolor{Gray}\textbf{51.7}} \\
        \bottomrule
    \end{tabular}
    }

\end{table*}

For MMDetection model parameter settings, training was conducted on models corresponding to three different input formats. The final selection of object detection models included Cascade R-CNN, PAA, RepPoints, and DINO. All models were built with ResNet101 as the backbone and employed the Progressive Asymmetric Feature Pyramid Network (PAFPN)~\cite{liu2018path} as the decoder. The input image resolution was set to 1280 × 1024 during training. Note that the dataset normalization parameters were calculated based on the statistics of the training set, rather than using zero mean and unit variance normalization as in the YOLO method. To enhance training stability and convergence speed, the AdamW optimizer was utilized in conjunction with the Scale strategy. All other hyperparameters remained consistent with the default settings of RepPoints.

For LoRAT~\cite{lin2024tracking} parameter settings, we opted for the Base version of the model and carefully adjusted several key hyperparameters to optimize performance. Specifically, we set the number of samples per epoch to 1800, the scale jitter factor to 0.2, and the minimum object size threshold to 1, while increasing the number of training epochs to 2000. Notably, larger versions of the model with more parameters did not yield any performance improvements. The remaining parameters were kept at their default values.

{\bf Inferencing Details.} During YOLO inference, the maximum number of detection boxes per image is set to 5, with both the IoU threshold and confidence threshold set to 0.2. For MMDetection inference, the maximum number of detection boxes per image is set to 3, while all other parameters remain unchanged. The detection results from all models are aggregated and processed using Weighted Box Fusion, where the IoU threshold is set to 0.5, the skip box confidence threshold is set to 0.3, and the fusion weight parameter is set to 0.1. The final detection boxes for each image are obtained through fusion, and the box with the highest confidence score is selected as the final result. If the highest confidence score is below 0.1, the image is considered to contain no target. To enhance tracking robustness, when the target position in the current frame deviates significantly from the previous frame (i.e., $S_t^{(i)}=0$ in Eq.~\ref{eq5}), the LoRAT tracking result is used as the final prediction. Note that since the target position in the first frame is not provided in Track 2, we did not use LoRAT in this track.

\begin{figure*}[t]
    \centering
    \includegraphics[width=\textwidth]{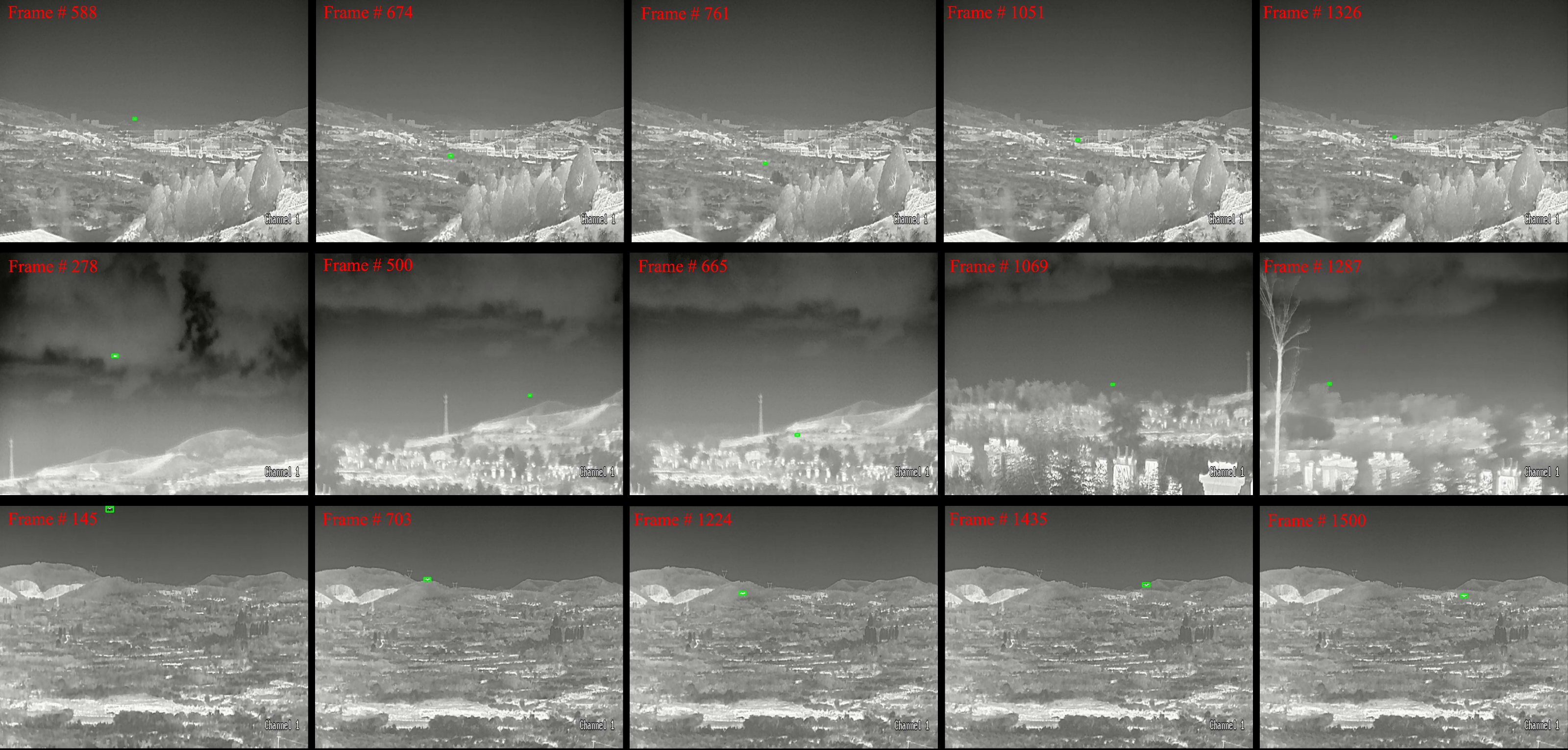} 
    \caption{The visualization of our method's inference results on the test set. Each row represents a different video sequence.}
    \label{fig:visresult_resized}
\end{figure*}

\begin{table}[t]
    \centering
    \caption{The relationship between the detection head, image size, and the scaling used in data augmentation, as well as their impact on detection performance.}
    \label{tab:ap50_combined}
    \definecolor{Gray}{gray}{0.9} 
    \resizebox{.48\textwidth}{!}{ 
    \begin{tabular}{c|cccc}
        \toprule
        \textbf{Methods} & \textbf{Head} & \textbf{Image Size} & \textbf{Scale} & \textbf{AP@50} \\
        \midrule
        \multirow{11}{*}{YOLOv11x-customized}  
            & P345  & 640×512  & 0.5  & 49.9 \\
            & P345  & 640×512  & 0.6  & 48.6 \\
            & P345  & 640×512  & 0.7  & 48.0 \\
            & P2345 & 640×512  & 0.5  & 53.4 \\
            & P2345 & 896×736  & 0.5  & 54.0 \\
            & P2345 & 1152×928 & 0.5  & 55.4 \\
            & P2345 & 1664×1376 & 0.5  & 56.3 \\
            & P2345 & 640×512  & 0.6  & 53.6 \\
            & \cellcolor{Gray}P2345 & \cellcolor{Gray}640×512  & \cellcolor{Gray}0.7  & \cellcolor{Gray}\textbf{56.4} \\
            & P2345 & 640×512  & 0.8  & 54.1 \\
            & P2345 & 640×512  & 0.9  & 54.7 \\
        \midrule
        \multirow{2}{*}{Cascade R-CNN}  
            & ---   & 640×512  & ---  & 52.4 \\
            & \cellcolor{Gray}---   & \cellcolor{Gray}1280×1024 & \cellcolor{Gray}---  & \cellcolor{Gray}\textbf{55.0} \\
        \midrule
        \multirow{2}{*}{RepPoints}     
            & ---   & 640×512  & ---  & 52.9 \\
            & \cellcolor{Gray}---   & \cellcolor{Gray}1280×1024 & \cellcolor{Gray}---  & \cellcolor{Gray}\textbf{56.5} \\
        \midrule
        \multirow{2}{*}{PAA}           
            & ---   & 640×512  & ---  & 52.5 \\
            & \cellcolor{Gray}---   & \cellcolor{Gray}1280×1024 & \cellcolor{Gray}---  & \cellcolor{Gray}\textbf{57.2} \\
        \midrule
        \multirow{2}{*}{DINO}         
            & ---   & 640×512  & ---  & 54.9 \\
            & \cellcolor{Gray}---   & \cellcolor{Gray}1280×1024 & \cellcolor{Gray}---  & \cellcolor{Gray}\textbf{58.3} \\
        \bottomrule
    \end{tabular}
    }
\end{table}

\subsection{Quantitative Evaluation}
 
Table~\ref{tab:main_results} displays the final scores of the top three teams in each track. We achieved first place in Track 1 and second in Track 2. Table~\ref{tab:map50_comparison} presents the detection performance of different methods under two frameworks. To enhance experimental efficiency, we utilized the training and validation sets from the 3rd Anti-UAV dataset, sampling 10\% of the images at regular intervals from each sequence. All models were trained on images at their original resolution. In the MMDetection framework, all models employed ResNet50 as the backbone, while in the YOLOv11 framework, the N model variant was used as the baseline. As shown in the table, Cascade R-CNN, PAA, RepPoints, and DINO demonstrated superior performance among the various MMDetection models, leading us to select these four models for the final ensemble. For the YOLOv11 framework, we explored multiple decoder and detection head designs tailored for small objects. The combination of the AFPN decoder with the $p2$ detection head yielded the most significant improvements, and further enhancements were achieved by integrating the C3K2 module into the detection head.

Table~\ref{tab:method_comparison} presents the detection and tracking metrics of the selected high-quality models on the four-fold dataset from the fourth competition dataset. Notably, all models in the MMDetection framework were trained on images with a resolution of 1280, while those in the YOLO framework were trained on images with a resolution of 640. Meanwhile, 20\% of the images in each sequence of the training set were sampled at fixed intervals for training. As shown in the tables, detection performance is generally positively correlated with tracking performance. Furthermore, using the frame difference of the first two frames as input before stitching the original images yields superior results. This finding indicates that leveraging temporal information for UAV detection can effectively suppress interference from complex backgrounds.

\begin{table}[ht]
    \centering
    \caption{The impact of the optimizer, batch normalization layer, and mixed-precision training on detection performance.}
    \label{tab:ap50_optimizers}
    \definecolor{Gray}{gray}{0.9} 
    \resizebox{.48\textwidth}{!}{ 
    \begin{tabular}{c|cccc} 
        \toprule
        \textbf{Methods} & \textbf{Optimizer}     & \textbf{BN} & \textbf{AMP} & \textbf{AP@50} \\
        \midrule
        \multirow{3}{*}{Cascade R-CNN} 
            & SGD           & BN     & \ding{55}  & 55.0 \\
            & AdamW+Scale   & SyncBN & \ding{51}  & 55.5 \\
            & \cellcolor{Gray}AdamW+Scale   & \cellcolor{Gray}SyncBN & \cellcolor{Gray}\ding{55}  & \cellcolor{Gray}\textbf{55.9} \\
        \midrule
        \multirow{3}{*}{PAA}          
            & SGD           & BN     & \ding{55}  & 57.2 \\
            & AdamW+Scale   & SyncBN & \ding{51}  & 58.5 \\
            & \cellcolor{Gray}AdamW+Scale   & \cellcolor{Gray}SyncBN & \cellcolor{Gray}\ding{55}  & \cellcolor{Gray}\textbf{59.4} \\
        \bottomrule
    \end{tabular}
    }
\end{table}

\subsection{Qualitative Evaluation}

Figure~\ref{fig:visresult_resized} presents the detection results of our method in four challenging scenarios, which primarily involve small object detection and interference from complex background noise. By integrating the temporal scale and incorporating the TC-filtering mechanism, our method effectively enhances the detection capability for small objects, ensuring high detection accuracy and robustness in complex environments. This improvement not only helps reduce the false detection rate but also enhances the model's adaptability across different scenarios.

\subsection{Ablation Study}

{\bf Impact of the Balance Between Image Scale and Detection Head on Detection Performance.} For YOLO-based models, the effectiveness of the $p2$ small-object detection head depends on target size, which is controlled by the scaling factor in data augmentation. Balancing these factors is crucial for optimal detection. As shown in Table~\ref{tab:ap50_combined}, increasing image resolution improves performance, but adjusting the scaling factor at the original resolution is even more effective. This ensures small object sizes align with the $p2$ head, maximizing its impact. For MMDetection-based models, a 2× image upscaling yields the best results, while 3× scaling degrades performance due to reduced batch size.

{\bf Impact of the Optimizer, Batch Normalization Layer, and Mixed-Precision Training on Detection Performance.} Table~\ref{tab:ap50_optimizers} presents the impact of the optimizer, BatchNorm type, and mixed-precision training (AMP) on detection performance. The results indicate that AdamW+Scale significantly outperforms SGD in terms of AP@50, suggesting that the combination of a more stable second-order optimization method and a scaling factor enhances gradient updates and improves model convergence. Additionally, SyncBN further boosts performance compared to standard BN, highlighting its ability to mitigate normalization bias and improve feature representation in small-batch training scenarios. Moreover, AMP does not provide performance gains in this experiment and may even introduce numerical precision errors that affect optimization stability. Overall, the combination of AdamW+Scale with SyncBN emerges as the optimal configuration, while the benefits of AMP require further evaluation based on specific hardware and dataset characteristics.

\begin{figure}[t]
  \centering

  \begin{subfigure}[b]{0.49\linewidth}
    \centering
    \includegraphics[width=\linewidth]{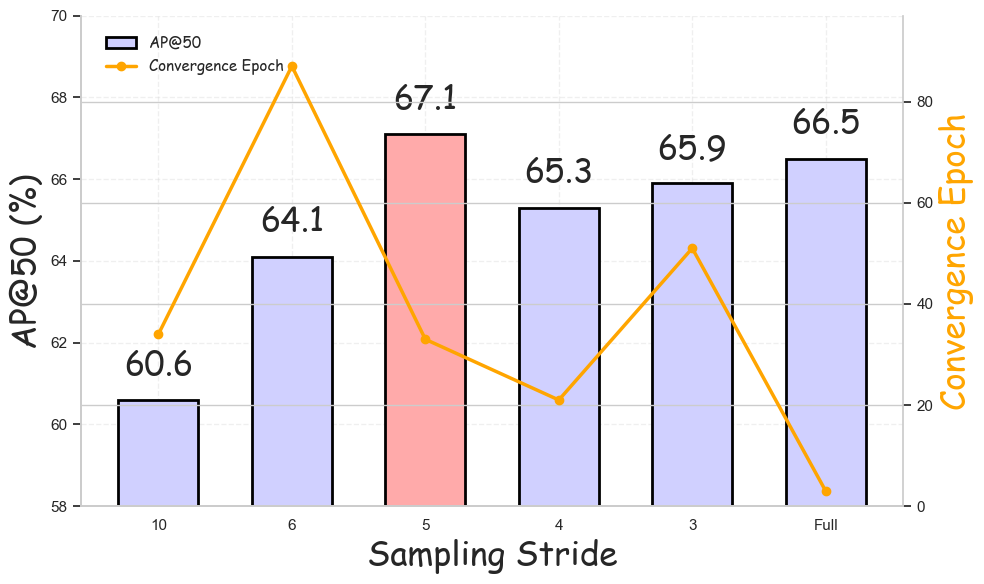}
    \caption{}
    \label{fig:sub-a}
  \end{subfigure}
  \hfill
  \begin{subfigure}[b]{0.49\linewidth}
    \centering
    \includegraphics[width=\linewidth]{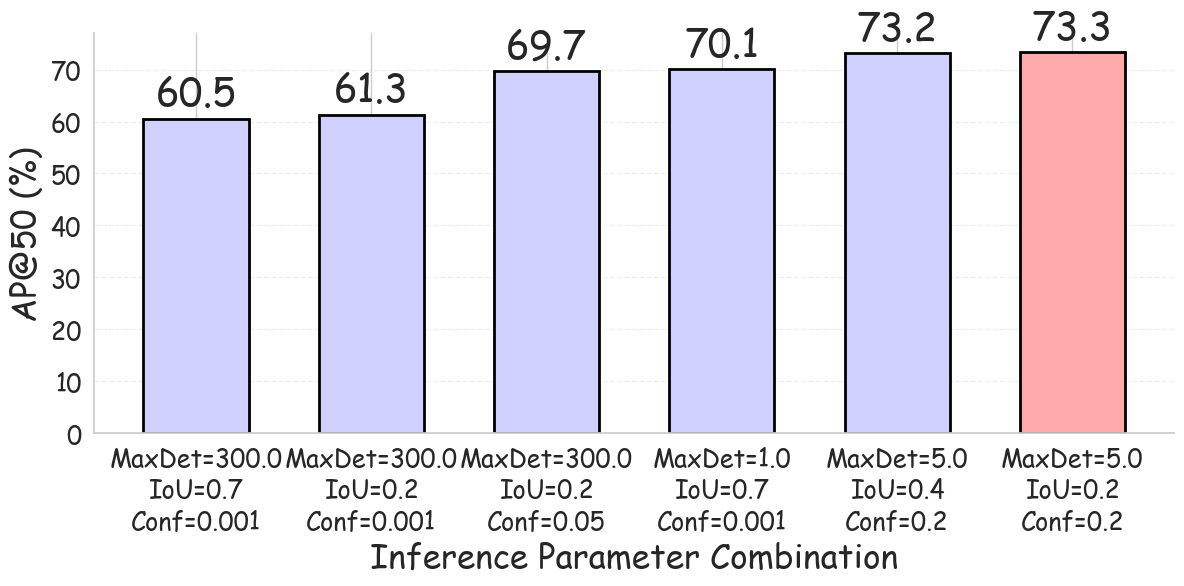}
    \caption{}
    \label{fig:sub-b}
  \end{subfigure}

  \caption{(a) Impact of the sampling stride on detection performance. (b) The impact of IoU threshold, confidence threshold, and the maximum number of detection boxes per image on detection performance.}
  \label{fig:combined}
\end{figure}

{\bf Impact of the Sampling Stride on Detection Performance.} Fig.~\ref{fig:sub-a} quantifies the impact of sampling stride on detection performance. Since the training data originates from video sequences, adjacent frames exhibit significant redundancy. Directly utilizing the full dataset may lead the model to overfit subtle inter-frame variations. Experimental results demonstrate that appropriately increasing the sampling stride effectively reduces redundancy while enhancing training efficiency. Notably, a stride of 5 achieves an AP@50 of 67.1\%, surpassing full-frame training at 66.5\%, indicating that a well-designed frame sampling strategy can enhance the model’s feature representation.

{\bf Impact of IoU Threshold, Confidence Threshold, and the Maximum Number of Detection Boxes per Image on Detection Performance.} As shown in Fig.~\ref{fig:sub-b}, in the single object detection scenario, lowering the IoU threshold to 0.2 improves recall, effectively adapting to the inherent blurriness of infrared targets. A confidence threshold of 0.2 achieves an optimal balance, while further reductions yield no significant gains. Reducing the maximum detections from 300 to 5 has minimal impact, confirming the redundancy of excessive bounding boxes in this context. The final configuration with an IoU threshold of 0.2, a confidence threshold of 0.2, and a maximum of 5 detections achieves the best trade-off between precision and recall.

\section{Conclusion}

In this paper, we propose an effective target tracking method that enhances performance by integrating global detection, motion-aware learning, and temporal priors. By concatenating the current frame with preceding frames, we transform a standard detection model into a robust tracking framework, leveraging temporal continuity to maintain high accuracy across frames. To further improve reliability, we introduce a prior assumption that the target is likely to appear near its previous location. Additionally, we propose an adaptive fusion strategy to integrate detection results from multiple models, enhancing overall performance. We achieve competitive performance in the Anti-UAV Challenge, securing 1st place in Track1 and 2nd place in Track2.

\section*{Acknowledgement}

This research was supported by the NSFC (62206134, 62361166670, 62176130, 62225604), the Fundamental Research Funds for the Central Universities 070-63233084, and the Shenzhen Science and Technology Program (JCYJ20240813114237048), Nankai International Advanced Research Institute (Shenzhen Futian), Nankai University. Computation is supported by the Supercomputing Center of Nankai University (NKSC). 

{\small
\bibliographystyle{ieee_fullname}
\bibliography{egbib}
}

\end{document}